\documentclass{article}

\PassOptionsToPackage{numbers, compress}{natbib}

\usepackage[preprint,final]{neurips_2026}

\usepackage[utf8]{inputenc} %
\usepackage[T1]{fontenc}    %
\usepackage{hyperref}       %
\usepackage{url}            %
\usepackage{booktabs}       %
\usepackage{amsfonts}       %
\usepackage{nicefrac}       %
\usepackage{microtype}      %
\usepackage{xcolor}         %

\usepackage{amssymb}            %
\usepackage{mathtools}          %
\usepackage{mathrsfs}           %
\usepackage{graphicx}           %
\usepackage{subcaption}         %
\usepackage[space]{grffile}     %
\usepackage{wrapfig}
\usepackage{enumitem}           %

\usepackage{tikz}
\newcommand{\colorcircle}[1]{%
  \protect\tikz[baseline=-0.6ex]\protect\draw[fill=#1, draw=#1] (0,0) circle (0.4em);%
}

\definecolor{mplcyan}{RGB}{86, 188, 190}
\definecolor{mplmagenta}{RGB}{170, 35, 180}

\usepackage{amsthm}

\newtheorem{lemma}{Lemma}

\usepackage[ruled]{algorithm2e}

\SetCommentSty{mycommfont}

\usepackage{xspace}
\usepackage{nimbusmono}

\usepackage{cleveref}

\usepackage{float}

\usepackage{amsmath,amsfonts,bm}

\def\eqref#1{equation~\ref{#1}}

\def\1{\bm{1}}

\def\rd{{\textnormal{d}}}

\def\vzero{{\bm{0}}}

\def\vmu{{\bm{\mu}}}
\def\vtheta{{\bm{\theta}}}
\def\va{{\bm{a}}}

\def\vf{{\bm{f}}}

\def\vl{{\bm{l}}}
\def\vm{{\bm{m}}}

\def\vp{{\bm{p}}}
\def\vq{{\bm{q}}}

\def\vs{{\bm{s}}}

\def\vw{{\bm{w}}}

\def\vz{{\bm{z}}}
\def\vtheta{{\bm{\theta}}}
\def\vphi{{\bm{\phi}}}
\def\vvarphi{{\boldsymbol\varphi}}

\def\mI{{\bm{I}}}

\def\mV{{\bm{V}}}
\def\mW{{\bm{W}}}

\def\mSigma{{\bm{\Sigma}}}

\DeclareMathAlphabet{\mathsfit}{\encodingdefault}{\sfdefault}{m}{sl}
\SetMathAlphabet{\mathsfit}{bold}{\encodingdefault}{\sfdefault}{bx}{n}

\def\gA{{\mathcal{A}}}
\def\gB{{\mathcal{B}}}

\def\gD{{\mathcal{D}}}

\def\gJ{{\mathcal{J}}}

\def\gL{{\mathcal{L}}}

\def\gN{{\mathcal{N}}}

\def\gS{{\mathcal{S}}}

\def\gU{{\mathcal{U}}}

\def\sR{{\mathbb{R}}}

\newcommand{\E}{\mathbb{E}}

\newcommand{\KL}{\mathbb{D}_{\mathrm{KL}}}

\DeclareMathOperator*{\argmax}{arg\,max}
\DeclareMathOperator*{\argmin}{arg\,min}

\newcommand{\sac}{\texttt{SAC}\xspace}
\newcommand{\xqc}{\texttt{XQC}\xspace}
\newcommand{\xqcfd}{\texttt{XQCfD}\xspace}
\newcommand{\mpo}{\texttt{MPO}\xspace}
\newcommand{\rlpd}{\texttt{RLPD}\xspace}

\newcommand{\redq}{\texttt{REDQ}\xspace}

\title{XQCfD: Accelerating Fast Actor-Critic Algorithms\\with Prior Data and Prior Policies}

\author{
Daniel Palenicek$^\star$\\
TU Darmstadt
\And
Florian Vogt$^\star$\\
KTH Royal Institute of Technology,
\And
Joe Watson$^\star$\\
University of Oxford\\
\AND
Ingmar Posner,\\
University of Oxford,\\
\And
Danica Kragic,\\
KTH Royal Institute of Technology
\And
Jan Peters\\
TU Darmstadt
}

\author{%
  Daniel Palenicek$^\star$$^{1,2}$\quad
  Florian Vogt$^\star$$^{3}$\quad
  Joe Watson$^\star$$^{4}$\quad
  Ingmar Posner$^{4}$\\
  \textbf{Danica Kragic}$^{3}$\quad
  \textbf{Jan Peters}$^{1,2,5,6}$ \\
  \thanks{$^\star$ denotes equal contribution.}
  $^{1}$Technical University of Darmstadt\; 
  $^{2}$hessian.AI\;
  $^{3}$KTH Royal Institute of Technology\; \\
  $^{4}$University of Oxford\;
  $^{5}$German Research Center for AI (DFKI)\\
  $^{6}$Robotics Institute Germany (RIG)\\
  \texttt{\{palenicek,joe\}@robot-learning.de}
}

\begin{document}

\maketitle

\begin{abstract}
For reinforcement learning in the real world, online exploration is expensive.
A common practice in robotic reinforcement learning is to incorporate additional data to improve sample efficiency.
Expert demonstration data is often crucial for solving hard exploration tasks with sparse rewards.
While prior data is used to augment experience and pre-train models, we show that the design of existing algorithms fails to achieve the sample efficiency that is possible in this setting due to a failure to use pretrained policies effectively.
We propose \xqcfd, which extends the sample-efficient \xqc actor-critic to learn from demonstrations,
using augmented replay buffers, pre-trained policies and stationary policy architectures, designed to avoid rapidly `unlearning' the strong initial policy like prior works.
We show our stationary network architecture enables policy improvement out-of-distribution better than standard network architectures due to its higher entropy predictions.
\xqcfd achieves state of the art performance across a range of complex manipulation tasks with sparse rewards from the popular Adroit, Robomimic and MimicGen benchmarks --- notably, with a low update-to-data ratio and no ensemble networks.
\end{abstract}

\section{Introduction}
\label{sec:intro}
Reinforcement learning (RL) provides a general framework for sequential decision-making, but high sample complexity remains the central barrier to real-world deployment~\citep{sutton1998introduction}.
Model-free deep actor-critic methods have made remarkable strides in sample efficiency through careful algorithmic and architectural choices: batch normalization, weight normalization, categorical critics, high update-to-data ratios, culminating in methods such as \texttt{CrossQ} \citep{bhatt2019crossq} and \xqc \citep{palenicek2025xqc, palenicek2025CrossQWN}.
In many practical settings, prior data is readily available before online interaction begins: expert demonstrations, previously collected offline datasets, or trajectories from related tasks.
Leveraging such data promises to accelerate online learning significantly.
Several research threads have combined RL with this additional data: \emph{offline-to-online RL} pre-trains a policy offline and finetunes it online; \emph{RL from demonstrations} (RLfD, \citep{vecerik2017leveraging,rajeswaran2018learning}) incorporates expert data into the agent's experience; \emph{RL from prior data}, e.g., RLPD \citep{ball2023efficient}, mixes sub-optimal offline and online data in the agent's replay buffer.
\emph{Imitation-bootstrapping} \citep{HuRSS24} interleaves on-policy and behavioral cloning actions during data collection to incorporate prior knowledge.

A recurring limitation in RLfD is failing to properly leverage the performance of \textit{behavioral cloning}~(BC \citep{Pomerleau1991}) at initialization
\citep{ball2023efficient,HuRSS24}.
Many current RLfD algorithms require millions of interactions to match BC performance, even though they eventually surpass it. 
This phenomenon occurs because the pre-trained policy is rapidly unlearned due to the learning dynamics of deep actor-critic algorithms and their use of function approximation.
The BC policy is not optimal for the randomly-initialized critic, and therefore the BC policy is rapidly unlearned in favor of a policy that is.
This phenomenon has been observed in inverse reinforcement learning, where it is more severe due to the additional learned reward \citep{orsini2021matters,watson2023coherent}.

We propose \xqcfd (\xqc from demonstrations), which extends the \xqc actor-critic implementation with a set of algorithmic changes to achieve state-of-the art sample efficiency for RLfD:
\begin{itemize}
    \item We perform policy pretraining on the demonstration data to start learning at the BC-level of performance, rather than learn from scratch.
    \item We also perform critic pretraining to ensure the initial actor and critic are coherent.
    \item To mitigate unlearning due to stochastic optimization, we use KL regularization between the policy and BC policy as an additional regularizer, instead of a maximum entropy one.
    \item To bridge KL and maximum entropy regularization, we adopt a `stationary' parametric policies which returns a constant, maximum entropy action distribution when out of distribution.
    This property enables exploration and policy improvement in states outside of the expert demonstration data, so the KL regularization does not limit policy improvement over BC.
\end{itemize}
While none of these design decisions are independently novel, this combination has not been used for RLfD, and we show empirically across a range of sparse manipulation tasks that this approach prevents BC unlearning and therefore enables dramatic sample efficiency compared to baselines when combined with XQC.  
In summary, our contributions are
\begin{enumerate}
    \item We combine the actor-critic algorithm \xqc with prior data and pre-trained policies for a sample-efficient RLfD algorithm, highlighting the crucial implementation details.
    \item We motivate and demonstrate the utility of stationary network architectures in RLfD.
    \item We provide thorough ablation experiments to validate our design decisions.
\end{enumerate}
\section{Related Work}
\label{sec:related_work}
Reinforcement learning from demonstrations or prior data is a long-established idea, especially for domains where exploration is challenging and expensive.
Currently, using prior data comes in three distinct settings, based on how the data is used and what is pre-trained.
\vspace{-0.5em}
\paragraph{Finetuning behavioral cloning with reinforcement learning.}
For challenging domains such as real-world robotics and strategy games, a practical solution for sample efficiency is to finetune BC policies 
\citep{schraudolph1993temporal,peters2006policy,kober2008policy,levine2016end,silver2016mastering}.
More modern practices combine an on- or off-policy deep RL algorithm with an additional BC-based loss or policy regularization in the actor objective to prevent unlearning
\citep{rajeswaran2018learning,rudner2021pathologies}.
We build on this work by incorporating a more sample-efficient off-policy algorithm that is able to finetune a BC policy without a significant performance drop, which is observed when using randomly initialized critics, e.g., the RLFT baseline in \citep{HuRSS24}.
\vspace{-0.5em}
\paragraph{Online reinforcement learning with additional data.}
With the advent of replay buffers in deep reinforcement learning, prior data can be easily integrated by merely augmenting the replay buffer.
Algorithm design decisions concern whether the replay buffer is simply augmented or multiplexed\footnote{Multiplexing refers to combining samples from several replay buffers within a single minibatch.}
\citep{vecerik2017leveraging,ball2023efficient}.
Imitation-bootstrapped RL \citep{HuRSS24} improves sample efficiency by incorporating BC action during interactions, and identifying dropout in the policy as an effective implementation detail and \redq \citep{chenrandomized} as an effective RL algorithm in this setting.
However, by not utilizing BC initialization effectively, buffer-based methods forfeit potentially significant sample efficiency gains, which we show in our experimental results.
\vspace{-0.5em}
\paragraph{Offline-to-online reinforcement learning.}
A related setting where the prior data is suboptimal considers continuing offline RL with online interactions and learning \citep{nair2020awac}.
This setting is challenging due to the difficulty of performing continual learning with function approximation with severe performance degradation, characterized as `catastrophic forgetting' or a lack of network `plasticity'.
Solutions include BC regularization \citep{zhaoadaptive}, careful sampling of offline and online samples from the replay buffer \citep{lee2022offline} and calibrating the $Q$ values during offline learning to mitigate policy unlearning during online interactions \citep{nakamoto2023cal}.
Our RLfD setting corresponds to offline-to-online RL with expert offline data, and we do not consider using suboptimal data as we desire strong initial BC policies.
\vspace{-0.5em}
\paragraph{Kullback-Leibler (KL)-regularized RL.}
While \xqc builds on \sac (soft actor-critic, \citep{haarnoja2018sac0}), \xqcfd uses a KL-regularized realization rather than a maximum entropy one, where the BC policy is a prior. 
\citet{siegelkeep2020} use the same regularization for offline RL, using \mpo \citep{abdolmaleki2018maximum} as the actor-critic algorithm.
\citet{rudner2021pathologies} use this regularization for RLfD with \sac, but use a nonparametric Gaussian process as the BC policy and a standard network architecture for the RL policy and minimizes their KL during pre-training. 
\citet{watson2023coherent} use this formulation with \sac, but in an inverse RL setting.

\section{Sample-Efficient Actor-Critic with Prior Data and Prior Policies}
\label{sec:method}
We consider the standard reinforcement learning setting, using a Markov decision process (MDP), defined by the tuple $(\gS, \gA, P, r, \gamma)$ where $\gS$ is the state space, $\gA$ is the action space, $P: \gS \times \gA \times \gS \to [0,1]$ are the transition dynamics, $r:\gS \times \gA \to \sR$ is the reward function, and ${\gamma \in [0,1)}$ is the discount factor.
At each timestep $h$, an agent observes state $\vs_h \in \gS$, selects action
${\va_h \sim \pi(\cdot \mid \vs_h)}$ according to a stochastic policy
$\pi: \gS \times \gA \to [0, 1]$, transitions to $\vs_{h+1} \sim P(\cdot \mid \vs_h, \va_h)$,
and receives reward $r_h = r(\vs_h, \va_h)$.
The objective is to find a policy $\pi^\star$ that maximizes the expected discounted return
$\gJ(\pi) = \E\!\left[\sum_{h=0}^{\infty} \gamma^h r_h\right]$.
A central quantity in reinforcement learning is the state-action value function
$Q^{\pi}(\vs, \va) = \E\!\left[\sum_{k=0}^{\infty} \gamma^k r_{h+k}
    \;\middle|\; \vs_h{\,=\,}\vs,\, \va_h{\,=\,}\va \right]$,
which satisfies the Bellman equation
$Q^{\pi}(\vs,\va) = r(\vs,\va) + \gamma\,\E_{\vs' \sim P(\cdot\mid\vs,\va),\, \va' \sim \pi(\cdot\mid\vs')}[Q^{\pi}(\vs',\va')]$.
Actor-critic algorithms maintain two function
approximators: an actor $\pi_{\vphi}$ parameterized by $\vphi$ that
represents the policy, and a critic $Q_{\vtheta}$ parameterized
by $\vtheta$ that estimates the value function.
Deep actor-critic algorithms maintain a replay buffer $\gB$ of past interactions. 
When learning from demonstrations, we also have access to a dataset of demonstrations $\gD$ of state-action trajectories capturing expert behavior.

\begin{figure}
    \vspace{-1em}
    \centering
    \includegraphics[width=\linewidth]{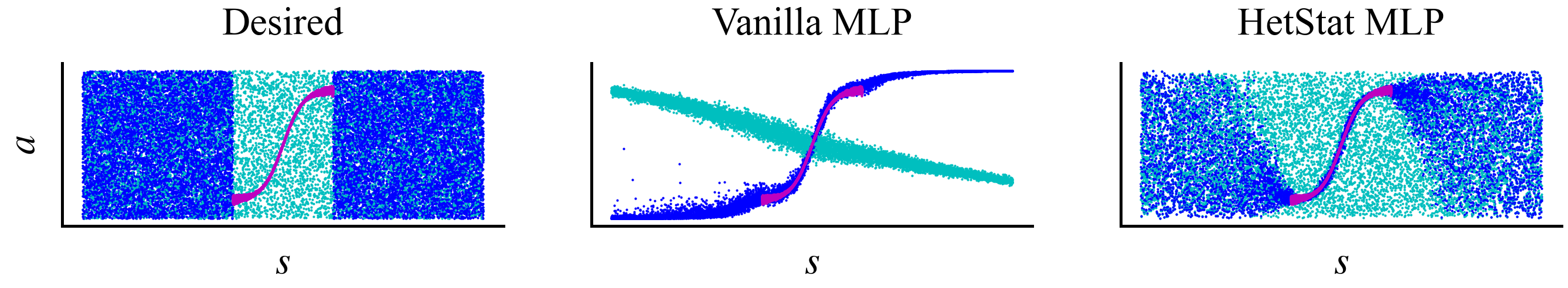}
    \caption{(Left) In the entropy-regularized RLfD setting, we desire a BC policy that fits the data (\colorcircle{mplmagenta}) while reflecting a maximum entropy prior out of distribution. (Centre) The typical multi-layer perceptron (MLP) architecture used in deep actor-critic methods does not have this property. 
    The out-of-distribution predictive behavior is undefined.
    Due to this behavior, KL regularization encourages the policy to stay close to suboptimal actions.
    (Right) By incorporating stationarity into the policy architecture, the policy is maximum entropy at initialization (\colorcircle{mplcyan}) and maximum entropy out-of-distribution after training (\colorcircle{blue}).
    In practice, there is a smooth transition between modeling the data and maximum entropy prior due to the smoothness of the function approximation. 
    }
    \label{fig:regression}
\end{figure}

Given a sample-efficient RL algorithm, incorporating prior expert data is a way of alleviating the pressure to explore and therefore achieve greater sample efficiency. 
While prior RLfD works adopt effective RL algorithms with $Q$ ensembles, layer norm, and high update-to-data ratios, the recently proposed \xqc \citep{palenicek2025xqc} has matched or outperformed these architectures w.r.t. parameter count, sample efficiency and wall-clock speed by combining batch norm \citep{ioffe2015batch}, weight norm \citep{lyle2024normalization} and a distributional critic \citep{bellemare2017c51}.
This combination ensures the critic's `effective' learning rate \citep{van2017l2} does not decrease and the critic's loss landscape is smooth and well-conditioned  \citep{santurkar2018does,ghorbani2019hessian_eigenspec}, which enables sample-efficient learning without any great computational overhead \citep{palenicek2025xqc}.
To leverage prior data, we
multiplex the replay buffers during online learning and pre-train the policy using BC.
The open research question concerns the tension between preventing unlearning and facilitating finetuning.
Can we finetune a parametric policy without unlearning the strong initialization?
To tackle this challenge, we use a policy architecture that has maximum entropy action predictions, as
we adopt the entropy-regularized RL formulation, which regularizes the policy against a prior distribution using the KL divergence.
While online RL uses a uniform action prior for exploration\footnote{A maximum entropy objective is equivalent to a KL minimization objective against a uniform distribution.} with temperature $\alpha \geq 0$, in the RLfD setting we replace this uninformative prior with the pre-trained BC policy $\pi_{\textsc{bc}}(\va\mid\vs)$, which also initializes $\pi_\vphi(\va\mid\vs)$, resulting in actor loss
\begin{align}
    \vphi_\star = \textstyle\argmax_\vphi \gL_\pi(\vphi) =
    \E_{\vs\sim\gB}\left[
    \E_{\va \sim \pi_\vphi(\cdot\mid\vs)}[Q_\vtheta(\vs,\va)]
    -
    \alpha \, \KL[\pi_\vphi(\va\mid\vs)\mid\mid    \pi_{\textsc{bc}}(\va\mid\vs)]
    \right].
    \label{eq:actor_obj}
\end{align}
Adopting the KL divergence provides strong policy improvement guarantees.
The solution to Equation \ref{eq:actor_obj} has monotonic improvement guarantees over the BC policy (Lemma \ref{lem:monotonic}).
\begin{lemma}{Monotonic policy improvement \citep{rawlik2013stochastic}.}\label{lem:monotonic}
Given a prior policy $p(\va{\mid}\vs)$, the pseudo-posterior policy update $\pi(\va{\mid}\vs)$ improves on the prior in a monotonic fashion for critic $Q$ due to the change-of-measure inequality \citep{donsker1983asymptotic},
\begin{align}
    \E_{\va\sim\pi(\cdot\mid\vs)}
    [Q(\vs,\va)]
    \geq
    \E_{\va\sim p(\cdot\mid\vs)}
    [Q(\vs,\va)]
    \;
    \forall \vs\in\gS,
    \;
    \pi(\va\mid\vs)
    \propto
    \exp\left(\frac{1}{\alpha} Q(\vs,\va)\right)\,p(\va\mid\vs).
\end{align}
\end{lemma}
However, in moving our objective from maximum entropy regularization with a uniform prior to KL regularization with the BC prior, we may have compromised our deep reinforcement learning algorithm.
In states outside of the demonstration distribution where we wish to perform online RL,  
$\pi_{\textsc{bc}}(\va\mid\vs)$ may be very different from the uniform prior $\gU(\va)$, especially as the predictive behavior of neural networks outside of the data distribution (OOD) is undefined, as illustrated in Figure \ref{fig:regression}. 
Therefore, during RL finetuning in out-of-distribution states, Equation \ref{eq:actor_obj} may encourage the agent to perform suboptimal actions rather than random exploration, and reducing $\alpha$ as a result also encourages unlearning the expert behavior.
To fix this issue, we desire a policy network that predicts a uniform action distribution when out of distribution.
Fortunately, this is possible by using policies with \emph{stationary} feature spaces.

\subsection{Maximum Entropy Policies using Stationary Features}
To finetune the BC policy in out-of-distribution states with maximum entropy RL using Equation \ref{eq:actor_obj}, we require a BC policy with the following characteristic,
\begin{align}
    \pi_{\textsc{bc}}(\va\mid\vs) = 
    \begin{cases}
    p_\gD(\va\mid\vs) \text{ if } \vs\in\gD, \text{ where $p_\gD$ is a conditional generative model of $\gD$},\\
    \gU(\va) \text{, the uniform prior, otherwise}.
    \end{cases}
\end{align}
A consequence of this characteristic is that, in the absence of any demonstration data, we desire a policy architecture where $\pi_{\textsc{bc}}(\va\mid\vs) = \gU(\va)\;\forall\; \vs\in\gS$.
Following soft actor-critic \citep{haarnoja2018sac0}, we use a Gaussian policy with a tanh transformation to bound the support of the action distribution to match the uniform prior. 
Therefore, this property is approximated by a latent action $\va=\tanh(\vz)$ with fixed zero-mean Gaussian predictions.
The desired out-of-distribution behavior described above can be achieved with a latent Gaussian model with stationary features $\vvarphi$.
Stationary features have constant self-similarity: the inner product $\varphi_\vtheta(\vs')^\top\varphi_\vtheta(\vs)$ is a function of a distance metric between $\vs'$ and $\vs$,
which means that the variance is a constant when $\vs'=\vs$, $p(z\mid\vs) = \gN(0, \sigma^2)$ as desired when $ p(\vw)=\gN(\vzero, \sigma^2\mI)$ due to the predictive distribution of Gaussian models shown below.
Our latent Gaussian policy is implemented using stationary last layer features, on top of MLP features, with Gaussian last layer weights $\vw$
such that 
$p(\vz\mid\vs) = \int p(\vz\mid\vs,\vw)\,p(\vw)\,\rd\vw$,
\begin{align*}
    z = \vw^\top\vvarphi_\vtheta(\vs),
    \quad
    p(\vw) = \gN(\vmu,\mSigma),
    \quad
    p(z\mid\vs) = 
    \gN(\vmu^\top\vvarphi_\vtheta(\vs), \vvarphi_\vtheta(\vs)^\top\mSigma\,\vvarphi_\vtheta(\vs)).  
    \label{eq:policy}
\end{align*}
This predictive behavior is preserved for out-of-distribution states after training, subject to the smoothness of the features. 
Unfortunately, this stationary property is only achieved exactly when $\varphi(\vs)$ is an infinitely large feature space and $\varphi_\vtheta(\vs)^\top\varphi_\vtheta(\vs')$ is expressed through its tractable kernel function $k(\vs,\vs')$ \citep{gpml}.
Fortunately, we can construct a \emph{finite} approximate to these feature spaces using \emph{random Fourier features} \citep{rahimi2007random} $\vf_k$ for kernel $k(\cdot,\cdot)$,
\begin{align}
    \vvarphi_{\vtheta,\mV}(\vs) = 
    \vf_k(\mV\vs),
    \quad
    V_{ij} \sim \gN(0, 1),
\end{align}
where $\vf_k$ are periodic functions required to approximate stationary kernel $k$ and $\mV$ is a fixed random weight matrix that constructs the random features  with $\vf_k$.
In practice, we apply the random features to neural network features rather than $\vs$ for greater function approximation capability, i.e., $\vf_k(\mV\varphi_{\text{\textsc{nn}}}(\vs))$, which preserves stationarity \citep{meronen2021periodic}.
We refer to this architecture as HetStat, as it is heteroscedastic\footnote{
Heteroscedastic predictions have input-varying variance, whereas homoscedastic predictions have constant variance.}
and stationary. 
HetStat policies were first proposed by \citet{watson2023coherent} in an inverse RL setting, combining it with SAC and the same actor objective (Equation \ref{eq:actor_obj}).
While this policy architecture was required for their learned reward, they also benefited from regularization properties of stationary priors observed by \citet{rudner2021pathologies}.
However, while \citet{rudner2021pathologies} used a (stationary) homoscedastic, nonparametric Gaussian process BC prior and an MLP policy, we and \citet{watson2023coherent} adopt a stationary network policy for both to benefit from parameter initialization and avoid the scaling issues that arise with nonparametric GPs \citep{gpml}. 

\begin{figure}[t]
\vspace{-1em}
\begin{algorithm}[H]
\KwData{Expert demonstrations $\gD$,
temperature $\alpha$,
parametric policy class $\pi_\vphi(\va\mid\vs)$,
parametric critic class $Q_\vtheta(\vs,a)$,
total steps $T$,
update-to-data ratio $N$,
policy delay $D$
}
\KwResult{$\pi_{\vphi_T}(\va\mid\vs)$, matching or improving the initial policy $\pi_{\vphi_1}(\va\mid\vs)$}
\tcp*[f]{pre-train policy and critic}\\
Train initial policy from demonstrations using Equation \ref{eq:bc_objective}, 
$\vphi_1 = \argmin_\vphi \gL_{\textsc{bc}}(\vphi).$\\
Pre-train critic on $\gD$ using $\vphi_1$ and Equation \ref{eq:critic_objective}.\\
\For(\tcp*[f]{finetune policy with reinforcement learning}){$h = 1 \rightarrow T$}
{
    Interact with environment
    $\vs_{h+1} \sim P(\cdot\mid\vs_h, \va_h)$,
    $\va_h \sim \pi_{\vphi_h}(\cdot\mid\vs_h)$, store in buffer $\gB$;\\
    \For{$n = 1 \rightarrow N$}
    {
        Update critic using $\gL_Q(\vtheta)$ (Equation \ref{eq:critic_objective}) on minibatch from $\gB$ and $\gD$; \\
        \If{$i\;\%\;D = 0$}
        {
            Update policy optimizing $\gL_\pi(\vphi)$ (Equation \ref{eq:actor_obj}) on minibatch from $\gB$ and $\gD$;
        }
    }
}
\caption{XQC from Demonstrations (\texttt{XQCfD}).}
\label{alg:xqcfd}
\end{algorithm}
\ifdefined\neurips
\vspace{-1.5em}
\fi
\end{figure}

\subsection{Off-policy reinforcement learning from demonstrations}
\label{sec:implementation}

We now outline the design of \xqc and our extensions for \xqcfd.

\textbf{Actor-critic algorithm.}
\xqc builds on SAC and adopts three design decisions: batch norm, weight norm, and a categorical critic.
Following \cite{bhatt2019crossq}, during critic learning we apply batch norm to the states and next states combined, i.e., computing $\vq, \vq' = Q_\vtheta([\vs, \vs'], [\va, \va'])$
to ensure the batch statistics correctly capture the state-action distribution seen by the critic.
In practice a target network is used to predict~$\vq'$.
Secondly, we apply weight norm to the critic network to constrain them to the unit sphere, i.e.,
$\tilde{\mW} = \mW / ||\mW||_2$.
This normalization keeps the effective learning rate constant \cite{lyle2024normalization}, which describes the phenomena where learning slows down as the weight norms of the parameters grow due to gradients scaling inversely proportional to the weights when batch norm is used~\citep{palenicek2025CrossQWN}.
The categorical critic has $101$ atoms and a cross entropy loss following the C51 architecture (Appendix \ref{sec:critic}, \citet{bellemare2017c51}).
For the actor, applying BN and WN is optional, but often beneficial and therefore used in our experiments.

For \xqcfd, we make three algorithmic changes: initializing the policy via pre-training, replacing the entropy term in the actor loss with KL regularization (Equation \ref{eq:actor_obj}), and maintaining both the demonstration and replay buffer separately.
The algorithm is summarized in Algorithm \ref{alg:xqcfd}.

\textbf{Model pre-training.}
For BC pre-training, we use a loss function that combines mean-squared error (MSE) and negative log likelihood (NLLH) metrics.
As our stochastic policy predicts both mean and variance, it is a heteroscedastic regression model. 
In deep learning, these models can suffer from ambiguity between signal and noise, resulting in regression fits that under-fits in the mean and overestimates the variance rather than accurately modeling complex functions.
The \textit{faithful} loss \citep{stirn2023faithful} resolves this issue by training the mean 
prediction $\vmu(\vs) = \vf_\vmu(\vvarphi(\vs))$ with the MSE objective, while training only 
the variance head with the NLLH.
Crucially, gradients are cut between the variance head and 
the shared feature space $\vvarphi(\vs)$, so that $\mSigma(\vs) = \vf_\mSigma(\mathrm{sg}(\vvarphi(\vs)))$, 
where $\mathrm{sg}(\cdot)$ denotes the stop-gradient operator, to ensure the features are optimized for the mean fit rather than uncertainty quantification.
The resulting loss is
\begin{align}
    \gL_\textsc{bc}(\vphi) &= \E_{\,\va,\vs\,\sim\gD}[\|\va - \vl(\vmu_\vphi(\vs))\|_2^2 - \log p(\vz\mid\vmu(\vs), \mSigma(\vs)^2)- \log|\nabla_\vz \vl(\vz)|],
    \label{eq:bc_objective}
\end{align}
where $\vl(\cdot)$ is the action transform, e.g., $\tanh$, of the latent Gaussian action $\vz\sim p(\cdot\mid\vmu,\mSigma)$.
We also pre-train the critic on the demonstration data using the same critic objective used for RL.

\textbf{Policy regularization.}
For the KL divergence between policies in Equation \ref{eq:actor_obj}, we use the closed-form KL between latent Gaussian actions, as the KL divergence between the tanh-squashed Gaussian has no closed-form expression and would require a Monte Carlo KL approximation.
Since the KL divergence is invariant under shared bijections \citep{polyanskiy2020infotheory_l1}, the closed-form Gaussian KL is in fact equal to our desired quantity.
Applying the closed-form KL to regularize predictions has also been used in supervised learning \citep{watson2020neural}. 
For the KL scaling term $\alpha$, we find a constant value is sufficient to regularize the update \citep{watson2023coherent} and did not jointly optimize the temperature like \citet{haarnoja2018sac0}.
The BC parameters initialize the RL policy, so the KL starts at zero.

\textbf{Experience replay.}
When combining data buffers, we combine the minibatches equally throughout training, coined as `symmetric sampling' in \citet{ball2023efficient}.
We also use the expert demonstration data to pre-compute the reward statistics for normalization and avoid online estimators.

\begin{figure}[!b]
    \centering
    \includegraphics[width=\textwidth]{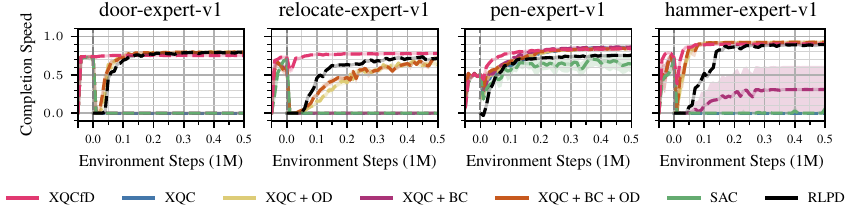}
    \vspace{-1em}
    \caption{Performance of \xqcfd and baselines on Adroit over 10 seeds, showing the IQM and 10th and 90th percentile stratified bootstrap confidence intervals.
    BC pretraining performance is shown before the vertical dashed line.
    Notably, in two of the tasks, BC achieves expert performance. 
    The benefit of improving upon BC is clear in this scenario, and baselines then require 100K--1M interactions to recover BC-level performance after unlearning.
    In the pen task, where BC is only $\sim$60\% successful, 
    \xqcfd only achieves a speed-up of 50K interactions, which supports our Hypothesis \ref{enum:sample}.
    }
    \label{fig:baseline_adroit}
\end{figure}

\section{Experiments}
\label{sec:experiments}

We now evaluate the empirical performance of \xqcfd across a diverse set of sparse-reward manipulation tasks drawn from three established benchmarks, comparing \xqcfd with HetStat and MLP policy architectures.
This section provides an overview of the experimental setup, including environments, implementation details and evaluation protocol. 
Section~\ref{sec:exp_results} presents our main results, comparing against relevant baselines in terms of sample efficiency and asymptotic performance. We then ablate key design decisions to isolate their individual contributions (Section~\ref{sec:exp_ablation}).

\paragraph{Environments.}
We evaluate \xqcfd on a total of $11$ sparse-reward manipulation tasks spanning $3$~popular benchmarks.
From Adroit~\citep{rajeswaran2018learning}, we use four dexterous hand tasks: \texttt{pen}, \texttt{door}, \texttt{hammer}, \texttt{relocate}.
From Robomimic~\citep{mandlekar2022matters}, we consider three tasks: \texttt{Lift}, \texttt{NutAssemblySquare}, and \texttt{PickPlaceCan}.
From MimicGen~\citep{mandlekar2023mimicgen}, we include four tasks: \texttt{StackThree}, \texttt{Coffee}, \texttt{HammerCleanup}, \texttt{MugCleanup}.
These environments are characterized by high-dimensional action spaces, contact-rich dynamics, and binary sparse rewards, making learning particularly challenging for agents relying solely on online exploration.
\vspace{-0.5em}
\paragraph{Demonstration data.}
For each task, we assume access to the standard dataset of expert demonstrations.
For most environments this is 200 demonstrations per task, apart from \texttt{StackThree\_D0} which provides 1000 demonstrations to variety of initial cube positions.
This data regime is reasonable to be collected manually by experts and provides performant BC policies.
We use the same expert datasets for BC pre-training, replay buffer augmentation and reward normalization statistics. 
\vspace{-0.5em}
\paragraph{Baselines.}
We compare \xqcfd against several baselines, including
BC \citep{Pomerleau1991},
\rlpd~\citep{ball2023efficient}, \xqc \citep{palenicek2025xqc}, and \sac \citep{haarnoja2018sac0}.
As a means of separating the contribution of the actor-critic algorithm from the RLfD, 
we consider three variants of \xqc, reflecting several RLfD approaches. These are (i) using BC pre-training (\xqc + BC), (ii) combining expert offline data with the online replay buffer (\xqc + OD) like \rlpd, and (iii) combining (i) and (ii) (\xqc + BC + OD).
For \sac, we use a HetStat policy.
\newpage
\textbf{Hypotheses.}
Our experimental design is built upon the following hypotheses.
\begin{enumerate}[label=\textcolor{blue}{\arabic*.}, ref=\arabic*]
    \item \xqcfd will improve upon BC performance without unlearning. \label{enum:unlearn}
    \item The sample efficiency of \xqcfd will depend on the performance of the BC policy.
    \label{enum:sample}
    \item \xqc will still outperform \sac and LayerNorm (\rlpd) in the sparse RLfD setting.\label{enum:xqc}
\end{enumerate}
If these three hypotheses are true, \xqcfd will be an effective RLfD implementation, subject to the quality of the demonstration data. 

\begin{figure}[!t]
    \vspace{-2em}
    \centering
    \includegraphics[width=\textwidth]{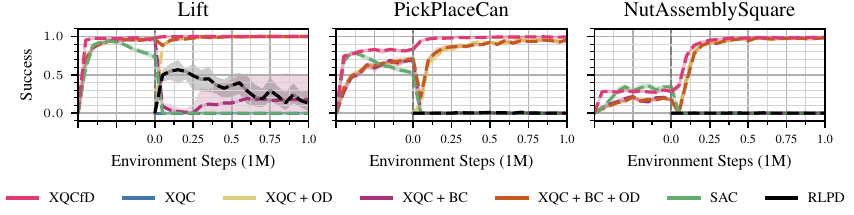}
    \vspace{-1em}
    \caption{Performance results on Robomimic over 10 seeds showing the IQM and 10th and 90th percentile stratified bootstrap confidence intervals.
    BC pretraining performance is shown before the vertical dashed line.
    OD refers to adding expert offline data.
    Robomimic includes three sparse manipulation tasks that are challenging to solve without leveraging demonstrations.
    For \texttt{Lift} and \texttt{PickPlaceCan}, the benefit of \xqcfd is clear, and BC performs close-to-expert performance and therefore \xqcfd is dramatically more stable than baselines. However, we see in the \texttt{Lift} task (simply lifting a block) \xqc + offline data learns extremely rapidly, and \xqc + offline data + BC unlearns very little. 
    However, for the harder \texttt{PickPlaceCan} task, unlearning is much more severe for our strong \xqc-based baselines and \xqcfd achieves a sample efficiency of 1M steps.
    For the harder \texttt{NutAssemblySquare} task, BC performs worse, therefore the performance gap between \xqcfd and \xqc + offline data is initially small and rapidly negligible, matching our Hypothesis \ref{enum:sample}.}
    \label{fig:baseline_robomimic}
    \vspace{-1em}
\end{figure}

\begin{figure}[!b]
    \centering
    \includegraphics[width=\textwidth]{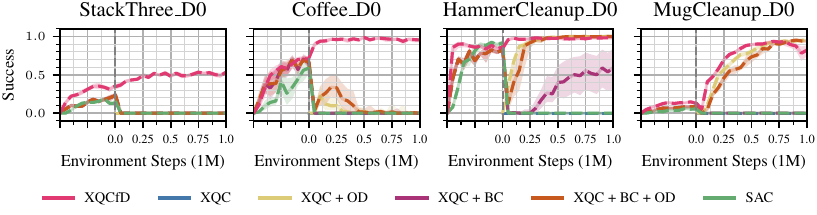}
    \vspace{-1em}
    \caption{Performance results of \xqcfd and baselines on MimicGen over 10 seeds, showing the IQM and 10th and 90th percentile stratified bootstrap confidence intervals.
    BC pretraining performance is shown before the vertical dashed line.
    OD refers to adding expert offline data.
    Compared to Robomimic, these tasks are slightly more complex due to a broader initial state distribution and longer task horizon. 
    For this suite, \xqcfd excels in tasks where the initial BC policy performs well. 
    In the Mug task, where the BC policy achieves a low success rate, \xqcfd is comparable to baselines without the benefit of pretraining. 
    For the hardest task, StackThree, \xqcfd is the only algorithm to achieve greater-than-zero success, but only improve the BC policy by ~10\%. 
    This suggests there are still improvements to be made before harder tasks can be mastered.
    }
    \vspace{-1em}
    \label{fig:baseline_mimicgen}
\end{figure}

\subsection{Experimental results}\label{sec:exp_results}

In Figures \ref{fig:baseline_adroit}, \ref{fig:baseline_robomimic},
\ref{fig:baseline_mimicgen}, we benchmark \xqcfd against baselines on Adroit, Robomimic and MimicGen respectively, looking at 11 tasks in total.
Across all tasks, Hypothesis \ref{enum:unlearn} holds, with little observable unlearning at the transition to BC. 
Improving on BC rather than starting from scratch yields notable gains in sample efficiency.
In \texttt{PickPlaceCan} the strongest baselines require $\sim$1M additional samples to reach \xqcfd performance and $\sim$500K in \texttt{relocate-expert-v1} and \texttt{HammerCleanup-D0}.
Moreover, \xqcfd is the only method to achieve non-zero success rate in the harder MimicGen tasks \texttt{StackThree-D0} and \texttt{Coffee-D0}.
For tasks \texttt{NutAssemblySquare} and \texttt{MugCleanup-D0}, the BC success rate was only in the region of 20-30\%, and there was a much smaller performance gap between \xqcfd and \xqc+ OD. 
We find a lower temperature was better in these environments, as the agent benefited from a greater trust region to improve on the worse-performing BC policy. 
However, this result confirms our Hypothesis \ref{enum:sample} that the value in \xqcfd relies on initial BC performance.
We found that this temperature parameter was also highly coupled to the unlearning phenomena, as \texttt{NutAssemblySquare} and \texttt{MugCleanup-D0} both exhibit a small drop in performance due to a lack of regularization. 
Figure \ref{fig:temperature_ablation} shows that a lower temperature can prevent premature convergence.

Regarding Hypothesis \ref{enum:xqc}, Figure \ref{fig:baseline_adroit} shows that while \sac and \rlpd can solve several tasks, similar \xqc variants outperform them for all but one environment. 
This result reflects previous findings on the qualities of \xqc \citep{palenicek2025CrossQWN,palenicek2025xqc}.
Moreover, \citet{ball2023efficient} tune the hyperparameters of \rlpd per environment. 
We were not able to tune \rlpd to solve Robomimic,
as observed by \citet{biza2025robot}.

\begin{figure}[t]
    \vspace{-2em}
    \centering
    \includegraphics[width=\linewidth]{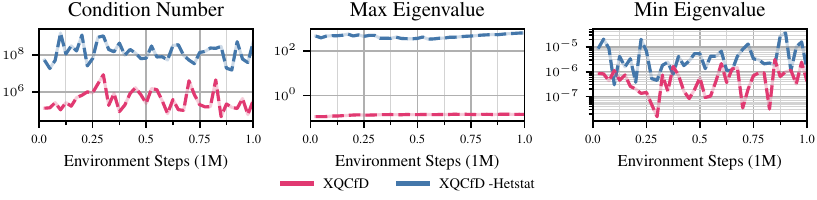}
    \caption{An empirical analysis of the optimization landscape of the actor with and without the HetStat policy for \texttt{PickPlaceCan}.
    The HetStat policy improves the condition number by around 100$\times$, which we attribute to the inherent regularization from the stationarity of the HetStat policy \cite{rudner2021pathologies}.
    The condition number analysis technique follows Behrooz et al. \cite{ghorbani2019hessian_eigenspec} and Palenicek et al. \cite{palenicek2025xqc}.
    }
    \label{fig:condition_number}
    \vspace{-1em}
\end{figure}

\subsection{Actor conditioning analysis}

\citet{palenicek2025xqc} have shown that critic architectures with better conditioning can be linked to improved RL training performance.
In Figure \ref{fig:condition_number}, we see that the HetStat policy improves the condition number of the actor loss by around 100$\times$, supporting the insight that stationary policies are provide a better loss landscape for learning due to their consistent predictions when out of distribution~\citep{rudner2021pathologies}.

\begin{figure}[b!]
    \vspace{-1em}
    \centering
    \includegraphics[trim={0 0.8cm 0 0},clip,width=\textwidth]{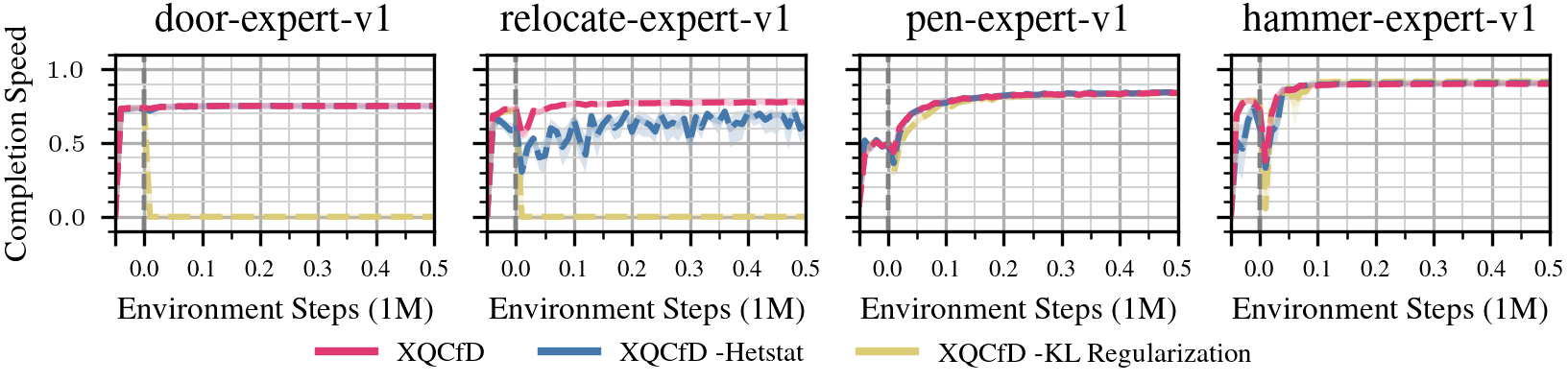}
    \includegraphics[trim={0 0.8cm 0 0},clip,width=\textwidth]{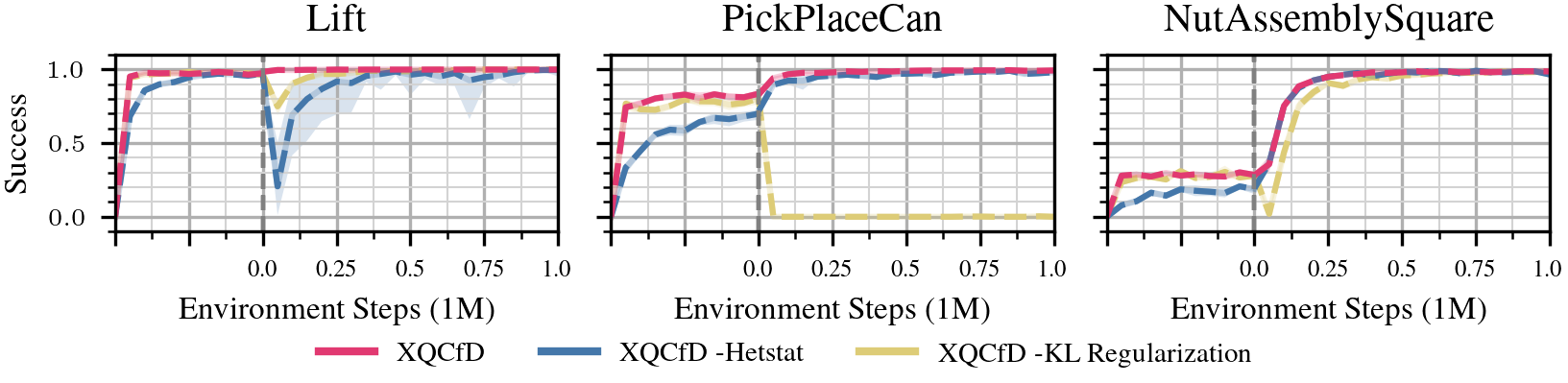}
    \includegraphics[width=\textwidth]{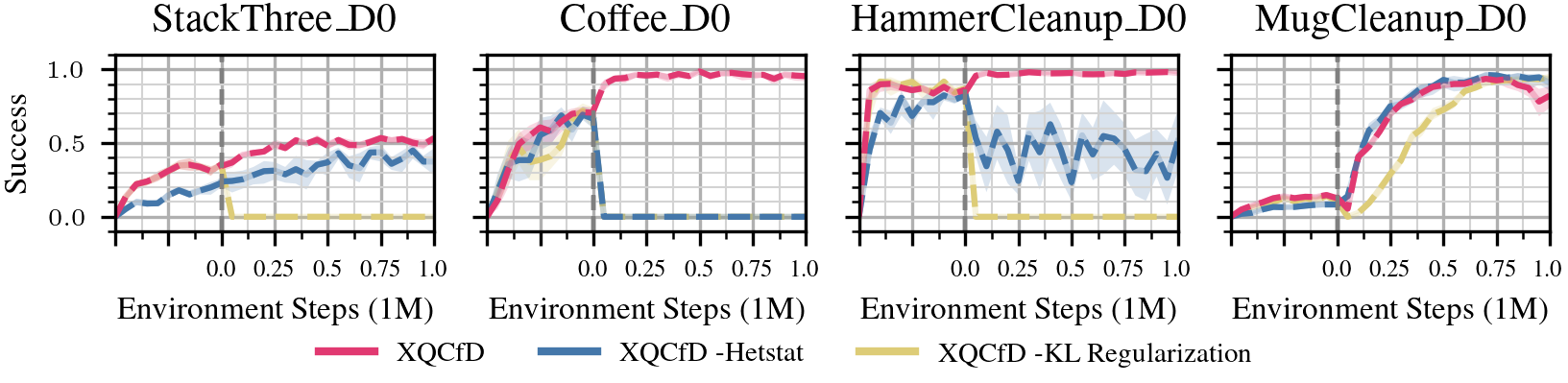}
    \caption{
    Ablation studies on \xqcfd, replacing the HetStat MLP with a standard MLP and replacing KL regularization with standard entropy regularization for the 
    Adroit (top), 
    Robomimic (middle) 
    and MimicGen (bottom) environments over 10 seeds, showing the IQM and 10th and 90th percentile stratified bootstrap confidence intervals.
    }
    \label{fig:adroit_ablation}
\end{figure}

\subsection{Ablation studies and hyperparameter sensitivity}\label{sec:exp_ablation}

In Figure \ref{fig:adroit_ablation}, 
we perform an ablation study on the HetStat policy and KL regularization.
For 6 out of 11 environments, the HetStat policy has clear benefits over the standard MLP for stable finetuning.

\begin{figure}[t]
    \vspace{-2em}
    \centering
    \includegraphics[width=\linewidth]{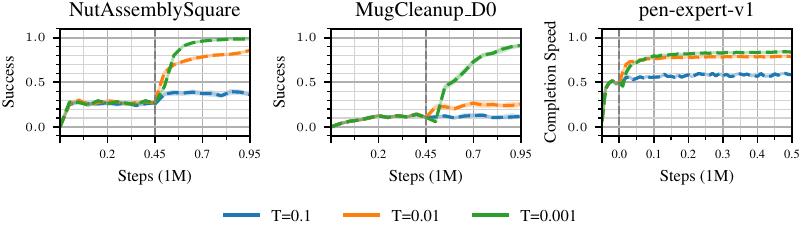}
    \caption{Sensitivity study of \xqcfd's temperature $\alpha$ that controls KL regularization against the BC policy.
    Evaluated over one task for Adroit, Robomimic and MimicGen over 10 seeds, showing the IQM and 10th and 90th percentile stratified bootstrap confidence intervals.
    Lower temperatures result in a larger performance drop on the transition from BC to RL due to unlearning, but lower temperatures also facilitate greater RL finetuning due to the smaller proximal regularization.}
    \label{fig:temperature_ablation}
\end{figure}

Regarding KL regularization, for 6 out of the 11 environments, removing KL regularization resulted in complete unlearning and failure to converge. 
For the 5 environments where KL regularization is not needed, these are environments where BC has a low success rate, so standard RL with auxiliary data is sufficient and unlearning does not need to be prevented.
While HetStat policies and KL regularization are needed under the same conditions, only one environment (\texttt{Coffee}) required both.
Our hypothesis is that the HetStat architecture is beneficial for stable finetuning, hence replacing it results in poor convergence (\texttt{relocate}, \texttt{Lift}, \texttt{PickPlaceCan}, 
\texttt{HammerCleanup}).
Removing KL regularization appears to frequently trigger rapid unlearning, although it is unclear why the agent cannot recover with additional experience, as the baselines appear to do when learning from scratch.
The baselines that do not have the additional data to aid exploration (\sac and \xqc+\texttt{BC}) consistently fail after unlearning.

In Figure \ref{fig:temperature_ablation}, we sweep across different values of the fixed temperature. Higher temperatures correspond to stronger BC regularization. When the BC policy has weak performance, online policy improvement struggles with stronger regularization, so
lowering the temperature increases the policy improvement during online training.

Lastly, in Figures \ref{fig:n_demos_adroit_ablation} to \ref{fig:n_demos_mimicgen_ablation} in Appendix \ref{sec:n_demos_ablation}, we provide a hyperparameter sensitivity experiment w.r.t. the number of expert demonstrations, which affects both the BC pretraining and auxiliary replay buffer.
The trend we observe is that if fewer demonstrations result in a weaker BC policy, RL finetuning can be weaker, which suggests the KL regularization should be tuned to account for a suboptimal BC policy when using fewer demonstrations.

\section{Conclusion}
\label{sec:conclusion}

We present \xqc from demonstrations (\xqcfd), an effective RLfD implementation that combines a sample-efficient RL algorithm, BC pretraining and regularization, and stationary policies.
\xqcfd solves an persistent limitation of prior deep off-policy RLfD approaches that are incapable of efficiently finetuning BC policies due to the randomly initialized critic causing the policy performance to drop below BC.
By finetuning BC policies instead of relearning from scratch, sample efficiency is significantly improved.
Across 11 sparse-reward manipulation tasks from Adroit, Robomimic and MimicGen, \xqcfd achieves state-of-the-art performance with a low update-to-data ratio and no Q-function ensembles. Our ablations confirm that our design decisions contribute to this performance in combination.
For future work, we plan to extend this algorithm to the action chunking setting with generative policies, which has been shown to improve BC performance on complex tasks \cite{chi2025diffusion} but is currently not stationary or straightforward to finetune with off-policy RL.

\textbf{Limitations.} Our experimental scope is restricted to expert demonstrations; we do not evaluate on suboptimal offline data, although \xqcfd could be applied to this setting with no adjustments.
Another current limitation is the fixed KL temperature, since RL performance can be sensitive to the BC quality when balancing regularization against premature convergence.
Future work could investigate adaptive temperature schedules that adjust regularization strength during training.

\begin{ack}
The work was enabled by the Berzelius resource provided by the Knut and Alice Wallenberg Foundation at the National Supercomputer Centre.
We further thank the Swedish Research Council and the Knut and Alice Wallenberg Foundation.
We gratefully acknowledge support from the hessian.AI Service Center (funded by the Federal Ministry of Education and Research, BMBF, grant no. 01IS22091), the hessian.AI Innovation Lab (funded by the Hessian Ministry for Digital Strategy and Innovation, grant no. S-DIW04/0013/003).
This research was funded by the research cluster “Third Wave of AI”, funded by the excellence program of the Hessian Ministry of Higher Education, Science, Research and the Arts, hessian.AI and has benefited from the early stages of the funding by the Deutsche Forschungsgemeinschaft (DFG, German Research Foundation) under Germany’s Excellence Strategy— EXC-3057; funding will begin in 2026.
This work was also supported by the Knut and Alice Wallenberg Foundation, Swedish Research Council and ERC AdV BIRD 88480, as well as the UKRI/EPSRC Programme Grant [EP/V000748/1].
Ingmar Posner holds concurrent appointments as a Professor of Applied AI at the University of Oxford and as an Amazon Scholar. This paper describes work performed at the University
of Oxford and is not associated with Amazon.
\end{ack}

\bibliographystyle{plainnat}
\bibliography{main}

\newpage
\appendix

\section{Ablation: Sensitivity to the Number of Expert Demonstrations}
\label{sec:n_demos_ablation}
\begin{figure}[H]
    \centering
    \includegraphics[width=\textwidth]{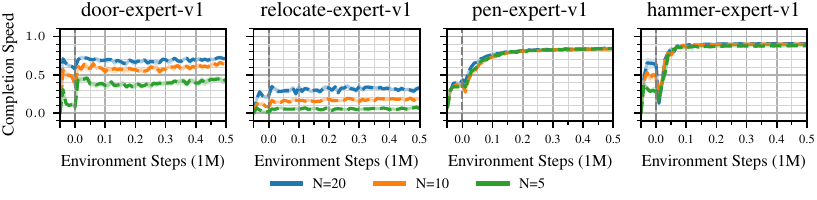}
    \caption{
    Sensitivity study on \xqcfd, replacing the number of expert demonstrations for the Adroit environments over 10 seeds,
    showing the IQM and 10th and 90th percentile stratified bootstrap confidence intervals.
    }
    \label{fig:n_demos_adroit_ablation}
\end{figure}

\begin{figure}[H]
    \centering
    \includegraphics[width=\textwidth]{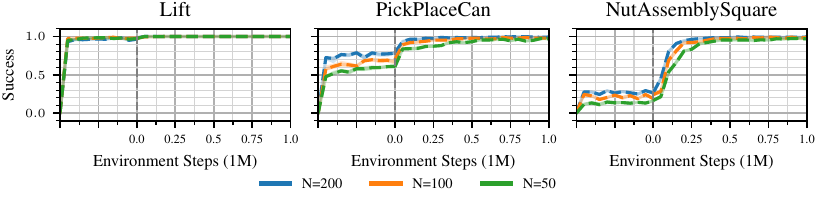}
    \caption{
    Sensitivity study on \xqcfd, replacing the number of expert demonstrations for the Robomimic environments over 10 seeds,
    showing the IQM and 10th and 90th percentile stratified bootstrap confidence intervals.
    }
    \label{fig:n_demos_robomimic_ablation}
\end{figure}

\begin{figure}[H]
    \centering
    \includegraphics[width=\textwidth]{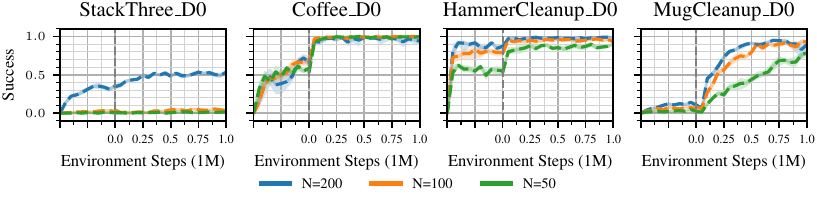}
    \caption{
    Sensitivity study on \xqcfd, replacing the number of expert demonstrations for the MimicGen environments over 10 seeds,
    showing the IQM and 10th and 90th percentile stratified bootstrap confidence intervals.
    }
    \label{fig:n_demos_mimicgen_ablation}
\end{figure}

\Cref{fig:n_demos_adroit_ablation,fig:n_demos_robomimic_ablation,fig:n_demos_mimicgen_ablation} show that \xqcfd remains effective across a range of demonstration dataset sizes on Adroit, Robomimic, and MimicGen.
As expected from Hypothesis~\ref{enum:sample}, gains scale with initial BC quality, indicating that the KL temperature can be tuned to extract further performance when fewer demonstrations are available.

\section{The Categorical Critic Objective}
\label{sec:critic}
In C51 \citep{bellemare2017c51}, the critic models the return distribution as a categorical distribution over $N$ fixed atoms $\{z_i\}_{i=1}^N$ with probabilities $\vp_{\vphi}(\vs, \va) = (p_{\vphi,1}(\vs,\va), \ldots, p_{\vphi,N}(\vs,\va))$. The target distribution is obtained by projecting the Bellman update $r + \gamma Z$ onto the atom support, yielding projected probabilities $\vm(\vs, \va, \vs')$. The critic is then trained by minimizing the cross-entropy loss,
\begin{align}
\gL_Q(\vtheta) = -\E_{\vs,\va,r,\vs' \sim \gB}\!\left[\sum_{i=1}^{N} m_i(\vs,\va,\vs')\log p_{\vtheta,i}(\vs,\va)\right],
\label{eq:critic_objective}
\intertext{
where the target projection coefficients are computed as
}
m_{i}(\vs,\va,\vs') = \sum_{j=1}^{N}\left[1 - \frac{\left|[\,r + \gamma z_j\,]_{z_1}^{z_N} - z_i\right|}{z_{i+1}-z_i}\right]_0^1 p_{\bar{\vtheta},j}(\vs',\va').
\end{align}
with $\va' \sim \pi_{\vphi}(\cdot\mid\vs')$.
Function $[\cdot]_a^b$ denoting clipping to $[a,b]$, and $\bar{\theta}$ denoting the parameters of a target network.

\section{Experimental Details}
\label{sec:experimant_details}

\paragraph{Evaluation protocol.}
All experiments were run using $10$ random seeds. For Robomimic and MimicGen, evaluations are performed every $50000$ environment steps, whereas for Adroit, evaluations are performed every $10000$ environment steps. In each evaluation phase, we execute $100$ rollouts using a deterministic policy. For Robomimic and MimicGen, we report the average success rate, and for Adroit, we report the completion speed, following the protocol of \citet{ball2023efficient}. 

\paragraph{Implementation details.}
Our implementation is based on the publicly available \xqc codebase~\citep{palenicek2025xqc}~(\href{https://github.com/danielpalenicek/xqc}{https://github.com/danielpalenicek/xqc}).
We use two sets of hyperparameters that differ only in their fixed temperature.
For the environments \texttt{MugCleanup}, \texttt{NutAssemblySquare}, \texttt{Pen}, and \texttt{Hammer}, we set the temperature to $0.001$, while for all other environments, we use a temperature of $0.01$.
We observe that stronger BC policies perform better with a higher temperature, whereas weaker BC policies benefit from a lower temperature.
Table \ref{hyperparameter} summarizes our hyperparameters alongside those used in the baseline methods.

\begin{table}[h]
\centering
\normalsize
\setlength{\tabcolsep}{6pt} %
\renewcommand{\arraystretch}{1.0} %
\begin{tabular}{lllll}
\toprule
\textbf{Hyperparameter} & \xqcfd & \xqc & \sac & \rlpd \\
\midrule
Learning rate & 3e-4 & 3e-4 & 3e-4 & 3e-4 \\
hidden dim & 512 & 512 & 512 & 256 \\
Policy Delay & 3 & 3 & 3 & 20 \\
Initial Temperature & 0.01 / 0.001 & 0.01 & 0.01 & 1.0 \\
Target entropy & - & $\left| A \right| / 2$ & $\left| A \right| / 2$ & $\left| A \right| / 2$ \\
Target network momentum & 0.005 & 0.005 & 0.005 & 0.005 \\
Update-to-data ratio & 2 & 2 & 2 & 20 \\
Critic ensemble size & 2 & 2 & 2 & 10 \\
Critic loss & Categorical & Categorical & MSE & MSE \\
Discount & Heuristic & Heuristic & Heuristic & 0.99 \\
\bottomrule
\end{tabular}
\caption{Overview of hyperparameters, including \sac (same parameters as \xqc) and critic loss settings.}
\label{hyperparameter}
\end{table}

RLPD uses environment-specific hyperparameters, which we summarize in Table~\ref{tab:rlpd_hyperparams}. Since RLPD does not provide hyperparameters for Robomimic, we manually adjusted them. We did not perform any additional tuning of these hyperparameters. 

\begin{table}[h]
\centering
\normalsize
\setlength{\tabcolsep}{6pt} %
\renewcommand{\arraystretch}{1.0} %
\begin{tabular}{lcccc}
\hline
\textbf{Environment} & \textbf{CDQ} & \textbf{Entropy} & \textbf{MLP Size} & \textbf{MLP Layers} \\ \hline
Adroit & True & False & 256 & 3 \\
Robomimic & True & False & 512 & 3 \\ \hline
\end{tabular}
\caption{Overview of RLPD environment-specific hyperparameters.}
\label{tab:rlpd_hyperparams}
\end{table}

\paragraph{Compute resources.}
The experiments for this paper were conducted on a high-performance computing cluster utilizing NVIDIA A100 and H100 GPUs.
A single GPU could run 5 seeds in parallel in ~16 hours, leveraging JAX vmap over seeds.
We estimate the total computational effort for this paper at approximately 1,000 GPU-hours. This total encompasses the primary experimental suite, ablation studies, hyperparameter tuning, and prior investigations.

\paragraph{Licenses.} We use the following assets:
XQC codebase~\citep{palenicek2025xqc} (MIT), 
Adroit~\citep{rajeswaran2018learning} (Apache 2.0), 
Robomimic~\citep{mandlekar2022matters} (MIT), 
MimicGen~\citep{mandlekar2023mimicgen} (NVIDIA License), 
RLPD~\citep{ball2023efficient} (MIT). 
All assets are used in accordance with their respective licenses.

\section{Discount Heuristic}
We set the discount factor based on the environment's horizon, following \citet{hansen2024tdmpc2scalablerobustworld}:
\[
\textstyle \gamma = \text{clip} \left( \frac{T/5 - 1}{T/5}, [0.95, 0.995] \right),
\]
where \(T\) denotes the episode length.

\newpage
\section*{NeurIPS Paper Checklist}

\begin{enumerate}

\item {\bf Claims}
    \item[] Question: Do the main claims made in the abstract and introduction accurately reflect the paper's contributions and scope?
    \item[] Answer: \answerYes{} %
    \item[] Justification: Yes, the abstract and introduction list the contributions of \xqcfd. We have additionally added a contributions paragraph to the end of the introduction.

\item {\bf Limitations}
    \item[] Question: Does the paper discuss the limitations of the work performed by the authors?
    \item[] Answer: \answerYes{}{} %
    \item[] Justification: We name limitations and assumptions in the paper and provide a dedicated limitations paragraph at the end of the conclusion.

\item {\bf Theory assumptions and proofs}
    \item[] Question: For each theoretical result, does the paper provide the full set of assumptions and a complete (and correct) proof?
    \item[] Answer: \answerYes{} %
    \item[] Justification: Our Lemma 1 is an established result we use to motivate our algorithm design.

    \item {\bf Experimental result reproducibility}
    \item[] Question: Does the paper fully disclose all the information needed to reproduce the main experimental results of the paper to the extent that it affects the main claims and/or conclusions of the paper (regardless of whether the code and data are provided or not)?
    \item[] Answer: \answerYes{} %
    \item[] Justification: We have taken great care to aid reproducibility. First, we descibte the proposed algorithm in detail throughout the main paper. Second, \Cref{sec:experimant_details} in the Appendix details implementation details, evaluation protocols and lists hyperparameters.

\item {\bf Open access to data and code}
    \item[] Question: Does the paper provide open access to the data and code, with sufficient instructions to faithfully reproduce the main experimental results, as described in supplemental material?
    \item[] Answer: \answerNo{} %
    \item[] Justification: We do not provide the code at the current time, however, we plan to release it upon acceptance of the paper.

\item {\bf Experimental setting/details}
    \item[] Question: Does the paper specify all the training and test details (e.g., data splits, hyperparameters, how they were chosen, type of optimizer) necessary to understand the results?
    \item[] Answer: \answerYes{} %
    \item[] Justification: We provide details in the main paper, as well as a dedicated section in the appendix listing experimental details, evaluation protocols and hyperparameters.

\item {\bf Experiment statistical significance}
    \item[] Question: Does the paper report error bars suitably and correctly defined or other appropriate information about the statistical significance of the experiments?
    \item[] Answer: \answerYes{} %
    \item[] Justification: We report IQM and 10th and 90th percentile stratified bootstrap confidence intervals across 10 random seeds for all training curves, following best practices in deep RL research (Agarwal et. al. 2021). 

\item {\bf Experiments compute resources}
    \item[] Question: For each experiment, does the paper provide sufficient information on the computer resources (type of compute workers, memory, time of execution) needed to reproduce the experiments?
    \item[] Answer: \answerYes{} %
    \item[] Justification: We include a paragraph in the Appendix, which details compute resources.
    
\item {\bf Code of ethics}
    \item[] Question: Does the research conducted in the paper conform, in every respect, with the NeurIPS Code of Ethics \url{https://neurips.cc/public/EthicsGuidelines}?
    \item[] Answer: \answerYes{} %
    \item[] Justification: We have read and conform to the code of ethics.

\item {\bf Broader impacts}
    \item[] Question: Does the paper discuss both potential positive societal impacts and negative societal impacts of the work performed?
    \item[] Answer: \answerNo{} %
    \item[] Justification: 
    This work concerns a sample-efficient deep actor-critic implementation and has no broader societal impact implications we believe would require a discussion.
    
\item {\bf Safeguards}
    \item[] Question: Does the paper describe safeguards that have been put in place for responsible release of data or models that have a high risk for misuse (e.g., pre-trained language models, image generators, or scraped datasets)?
    \item[] Answer: \answerNA{} %
    \item[] Justification: Our work focuses on fundamental research on small scale deep reinforcement learning algorithms. We neither train, nor provide any large scale model checkpoints that could be misused. Therefore, this question is not applicable.

\item {\bf Licenses for existing assets}
    \item[] Question: Are the creators or original owners of assets (e.g., code, data, models), used in the paper, properly credited and are the license and terms of use explicitly mentioned and properly respected?
    \item[] Answer: \answerYes{} %
    \item[] Justification: We cite used codebases and benchmarks extensively in the paper and have added a section with references and corresponding licenses to the Appendix.

\item {\bf New assets}
    \item[] Question: Are new assets introduced in the paper well documented and is the documentation provided alongside the assets?
    \item[] Answer: \answerNA{} %
    \item[] Justification: This paper does not release new assets.

\item {\bf Crowdsourcing and research with human subjects}
    \item[] Question: For crowdsourcing experiments and research with human subjects, does the paper include the full text of instructions given to participants and screenshots, if applicable, as well as details about compensation (if any)? 
    \item[] Answer: \answerNA{} %
    \item[] Justification: This work does not involve crowdsourcing nor research with human subjects.

\item {\bf Institutional review board (IRB) approvals or equivalent for research with human subjects}
    \item[] Question: Does the paper describe potential risks incurred by study participants, whether such risks were disclosed to the subjects, and whether Institutional Review Board (IRB) approvals (or an equivalent approval/review based on the requirements of your country or institution) were obtained?
    \item[] Answer: \answerNA{} %
    \item[] Justification: This research does not involve crowdsourcing nor research with human subjects.

\item {\bf Declaration of LLM usage}
    \item[] Question: Does the paper describe the usage of LLMs if it is an important, original, or non-standard component of the core methods in this research? Note that if the LLM is used only for writing, editing, or formatting purposes and does \emph{not} impact the core methodology, scientific rigor, or originality of the research, declaration is not required.
    \item[] Answer: \answerNo{} %
    \item[] Justification: This work does not use LLMs as part of the core methodology.

\end{enumerate}

\end{document}